\documentclass[11pt,letterpaper]{article}
\usepackage[letterpaper]{geometry}
\usepackage{acl2012}
\usepackage{times}
\usepackage{graphicx} 
\usepackage{amssymb}
\usepackage{latexsym}
\usepackage{amsmath}
\usepackage{multirow}
\usepackage{url}
\usepackage{amsfonts}
\usepackage{dsfont}
\makeatletter
\newcommand{\@BIBLABEL}{\@emptybiblabel}
\newcommand{\@emptybiblabel}[1]{}
\makeatother
\usepackage[hidelinks]{hyperref}

\newcommand{\change}[1]{}

\title{Deep Recurrent Models with Fast-Forward Connections for Neural Machine Translation}

\author{Jie Zhou \ Ying Cao \ Xuguang Wang \ Peng Li \ Wei Xu \\
Baidu Research - Institute of Deep Learning \\
Baidu Inc., Beijing, China \\
{\tt \{zhoujie01,caoying03,wangxuguang,lipeng17,wei.xu\}@baidu.com} \\
}

\date{}

\begin{document}
\maketitle
\begin{abstract}
  Neural machine translation (NMT) aims at solving machine translation (MT) \mbox{problems} using neural networks and has exhibited
  promising results in recent years.  \mbox{However}, most of the existing NMT models are shallow and there is still a performance gap
  between a \mbox{single} NMT model and the best \mbox{conventional} MT system.   In this work, we introduce a new type of linear
  connections, named fast-forward connections, based on deep Long Short-Term Memory (\mbox{LSTM}) networks, and an interleaved
  bi-directional architecture for stacking the LSTM layers.  Fast-forward connections play an essential role in propagating the
  gradients and building a deep topology of depth $16$. On the WMT'14 English-to-French task, we \mbox{achieve} BLEU=$37.7$ with a
  single attention model, which outperforms the corresponding single shallow model by $6.2$ BLEU points. This is the first time that a
  single NMT model achieves state-of-the-art performance and  outperforms the best conventional model by $0.7$ BLEU points. We can still
  achieve BLEU=$36.3$ even without using an attention mechanism. After  special handling of unknown words and  model
  ensembling,  we obtain the best score reported to date \change{Is it the best score every reported? You can say "obtain the best score reported to
  date"} on this task with BLEU=$40.4$. Our models are also  validated  on the more difficult WMT'14 English-to-German task.
\end{abstract}

\section{Introduction}

Neural machine translation (NMT) has attracted a lot of interest in solving the machine translation (MT) problem in recent
years ~\cite{Kalchbrenner-Blunsom-EMNLP2013,Sutskever-Le-NIPS2014,Bahdanau-Bengio-ICLR2015}. Unlike conventional statistical machine
translation (SMT) systems ~\cite{Koehn-Marcu-NAACL2003,Durrani-Heafield-WMT2014} which consist of multiple separately tuned components, NMT
models encode the source sequence into continuous representation \mbox{space} and generate the target sequence in an end-to-end
fashon. Moreover, NMT models can \mbox{also} be easily adapted to other tasks such as dialog
\mbox{systems}~\cite{Vinyals-Le-ICML2015}, question answering systems~\cite{Yu-Zhou-ARXIV2015} and  image caption
generation~\cite{Mao-Yuille-ICLR2015}.

In general, there are two types of NMT topologies: the encoder-decoder network~\cite{Sutskever-Le-NIPS2014} and the
attention network~\cite{Bahdanau-Bengio-ICLR2015}. The encoder-decoder network represents the source sequence with a fixed
dimensional vector and the target sequence is generated from this vector  word by word. The attention network \mbox{uses}
the representations from all time steps of the input sequence to build a detailed relationship between the target words and the input
words. Recent results show that the systems based on these models can achieve similar performance to conventional SMT
\mbox{systems}~\cite{Luong-Zaremba-ACL2015,Jean-Bengio-ACL2015}.

However,  a single neural model of either of the above types has not been competitive with the best conventional
system~\cite{Durrani-Heafield-WMT2014} when \mbox{evaluated} on the WMT'14 English-to-French task. The best BLEU score from a single
model with six layers is only  $31.5$~\cite{Luong-Zaremba-ACL2015} while the conventional method of ~\cite{Durrani-Heafield-WMT2014}
achieves $37.0$.

We focus on improving the single model performance by increasing the model depth. Deep topology has been proven to outperform the shallow
architecture in computer vision. In the past two years  the top positions of the ImageNet contest have always been occupied
by systems with \change{was:more than} tens or even hundreds of layers~\cite{Szegedy-Rabinovich-CVPR2015,He-Sun-ARXIV2015}. But in
NMT, the biggest depth used successfully is \mbox{only} six~\cite{Luong-Zaremba-ACL2015}. We attribute this problem to the properties of
the Long Short-Term Memory (\mbox{LSTM})~\cite{Hochreiter-Schmidhuber-NC1997} which is widely used in NMT. In the LSTM,
there are more non-linear activations than in convolution layers. These activations significantly decrease the  magnitude of the
gradient in the deep topology, especially when the gradient propagates in recurrent form. There are also many efforts to increase the depth of
the LSTM such as the work by ~\newcite{Kalchbrenner-Graves-Arxiv2015}, where the shortcuts do not avoid the nonlinear and
recurrent computation.

In this work, we introduce a new type of linear connections for multi-layer recurrent networks. These connections, which are called
fast-forward connections, play an essential role in building a deep topology with depth of $16$.   In addition, we introduce an interleaved bi-directional architecture  to stack LSTM layers in the
encoder.  This topology can be used for both the encoder-decoder network and the attention  network. On
the WMT'14 English-to-French task, this is the deepest NMT topology that has ever been investigated. With our deep attention model,
the BLEU score can be improved to  $37.7$  outperforming the shallow model which has six layers~\cite{Luong-Zaremba-ACL2015} by $6.2$
BLEU points. This is also the first time on this task that a single NMT model achieves state-of-the-art performance and outperforms the best
conventional SMT system~\cite{Durrani-Heafield-WMT2014} with an improvement of $0.7$. Even without using the attention mechanism, we
can still achieve $36.3$ with a single model.    After model ensembling and unknown word processing, the BLEU score can be further improved to
$40.4$. When evaluated on the subset of the test corpus without unknown words, our model achieves $41.4$. As a reference, previous
work showed that oracle re-scoring of the $1000$-best sequences  generated by the SMT model can achieve the BLEU score of about
$45$~\cite{Sutskever-Le-NIPS2014}. Our models are also validated on the more difficult WMT'14 English-to-German task.

\section{Neural Machine Translation}
\label{sec:NMT}
Neural machine translation aims at generating the target word sequence $\boldsymbol{y}=\{y_1,\dots,y_n\}$ given the source word sequence
$\boldsymbol{x}=\{x_1,\dots,x_m\}$ with neural models. In this task, the likelihood $p(\boldsymbol{y}\mid\boldsymbol{x},\boldsymbol{\theta})$
of the target sequence will be maximized ~\cite{Forcada-Neco-1997} with parameter $\boldsymbol{\theta}$ to learn:
\begin{equation}
\label{eqn:NMT-Model-Likelihood}
  p (\boldsymbol{y}\mid\boldsymbol{x};\boldsymbol{\theta}) = \prod_{j=1}^{m+1}
  p(y_j\mid\boldsymbol{y}_{0:j-1},\boldsymbol{x};\boldsymbol{\theta})
\end{equation}
where $\boldsymbol{y}_{0:j-1}$ is the sub sequence from $y_0$ to $y_{j-1}$. $y_0$ and $y_{m+1}$ denote the start mark and end mark of target
sequence respectively.

The process can be explicitly split into an encoding part, a decoding part and the interface between these two parts. In the
encoding part, the source sequence is processed and transformed into a group of vectors $\boldsymbol{e}=\{e_1,\cdots,e_m$\} for each time
step. Further operations will be used at the interface part to extract the final representation $\boldsymbol{c}$ of the source sequence from
$\boldsymbol{e}$. At  the decoding step, the target sequence is generated from the representation $\boldsymbol{c}$.

Recently, there have been two types of NMT models which are  \mbox{different} in the interface part. In the encoder-decoder
model~\cite{Sutskever-Le-NIPS2014},  a single vector extracted from $\boldsymbol{e}$ is used as the representation. In the attention
model~\cite{Bahdanau-Bengio-ICLR2015}, $\boldsymbol{c}$ is dynamically obtained according to the relationship between the target sequence and
the source sequence.

The recurrent neural network (RNN), or its specific form the LSTM, is generally used as the basic
\mbox{unit} of the encoding and decoding part. However, the topology of most of the existing models is shallow. In the
attention network, the encoding part and the decoding part have only one LSTM layer respectively.  In the
encoder-decoder network, researchers have used at most six LSTM layers~\cite{Luong-Zaremba-ACL2015}. Because machine
translation is a  difficult problem, we believe more complex encoding and decoding architecture  is needed for modeling the relationship
between the source sequence and the target sequence. In this work, we focus on enhancing the complexity of the encoding/decoding architecture
by increasing the model depth.

Deep neural models have been studied in a wide range of problems. In computer vision, models with more than ten\change{was:tens "more than
tens" would mean at least a hundred} convolution layers outperform  shallow ones on a series of image tasks in \mbox{recent}
years~\cite{Srivastava-Schmidhuber-Arxiv2015,He-Sun-ARXIV2015,Szegedy-Rabinovich-CVPR2015}. Different kinds of shortcut connections are
proposed to decrease the length of the gradient propagation path. Training networks based on \mbox{LSTM} layers, which
are widely used in language problems, is a much more challenging task. Because of  the existence of
many more nonlinear activations and the recurrent computation, gradient values are not stable and are generally
smaller. Following the same spirit for convolutional networks, a lot of effort has also been spent
on training deep LSTM networks.  Yao et al.~\shortcite{Yao-Dyer-ARXIV2015} introduced depth-gated shortcuts,
connecting LSTM cells at adjacent layers, to provide a fast way to propagate the gradients. They validated the modification
of these shortcuts on an MT task and a language modeling task. However, the best \mbox{score} was obtained
using models with three layers. Similarly, Kalchbrenner et al.~\shortcite{Kalchbrenner-Graves-Arxiv2015} proposed a two dimensional
structure for the LSTM. Their structure decreases the number of nonlinear activations and path length. However, the gradient
propagation still relies on the recurrent computation. The investigations were also made on question-answering to encode the questions, where
at most \mbox{two} LSTM layers were stacked~\cite{Hermann-Blunsom-Arxiv2015}.

Based on the above considerations, we propose new connections to facilitate gradient propagation in the following section.

\section{Deep Topology}

We build the deep LSTM network with the new proposed linear connections. The shortest paths through the proposed connections do not include
any nonlinear transformations and do not rely on any recurrent computation. We call these connections fast-forward connections.  Within the
deep topology, we also introduce an interleaved bi-directional architecture to stack the LSTM layers.

\subsection{Network}

\begin{figure}[!ht]
  \begin{center}
  \includegraphics[angle=0,width=0.9\columnwidth]{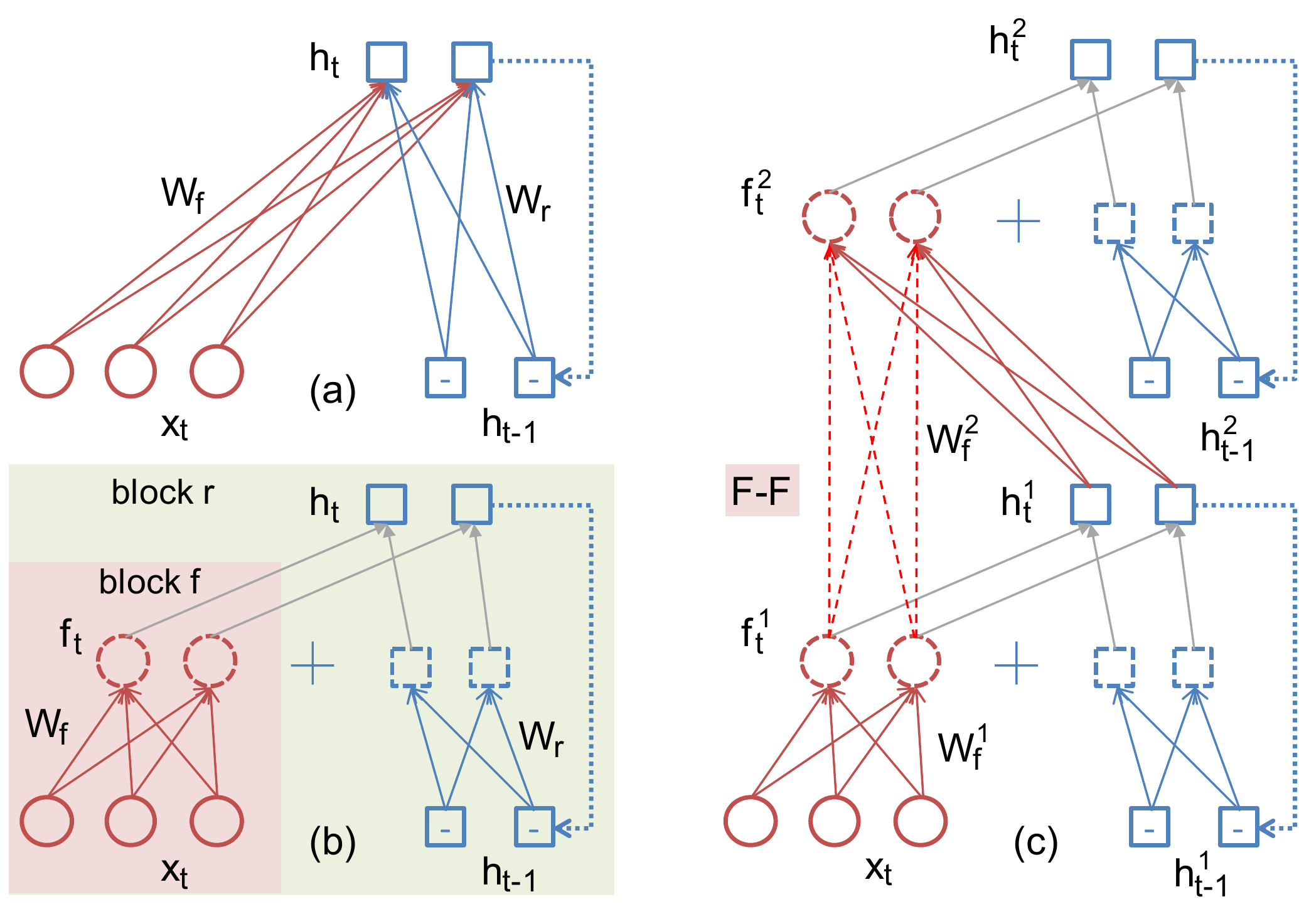}
  \caption{RNN models. The recurrent use of a hidden state is denoted by dotted lines. A ``-'' mark denotes the hidden value
  of the previous time step. (a): Basic RNN. (b): Basic RNN with intermediate computational state and the sum operation (+) followed
  by activation. It consists of block ``f'' and block ``r'',  and is equivalent to (a).  (c):Two stacked RNN layers with F-F connections
  denoted by dashed red lines.}
  \label{fig:NMT-Model-RNN-FF}
  \end{center}
\end{figure}

\begin{figure*}[!htbp]
\begin{center}
\includegraphics[angle=0,width=1.9\columnwidth]{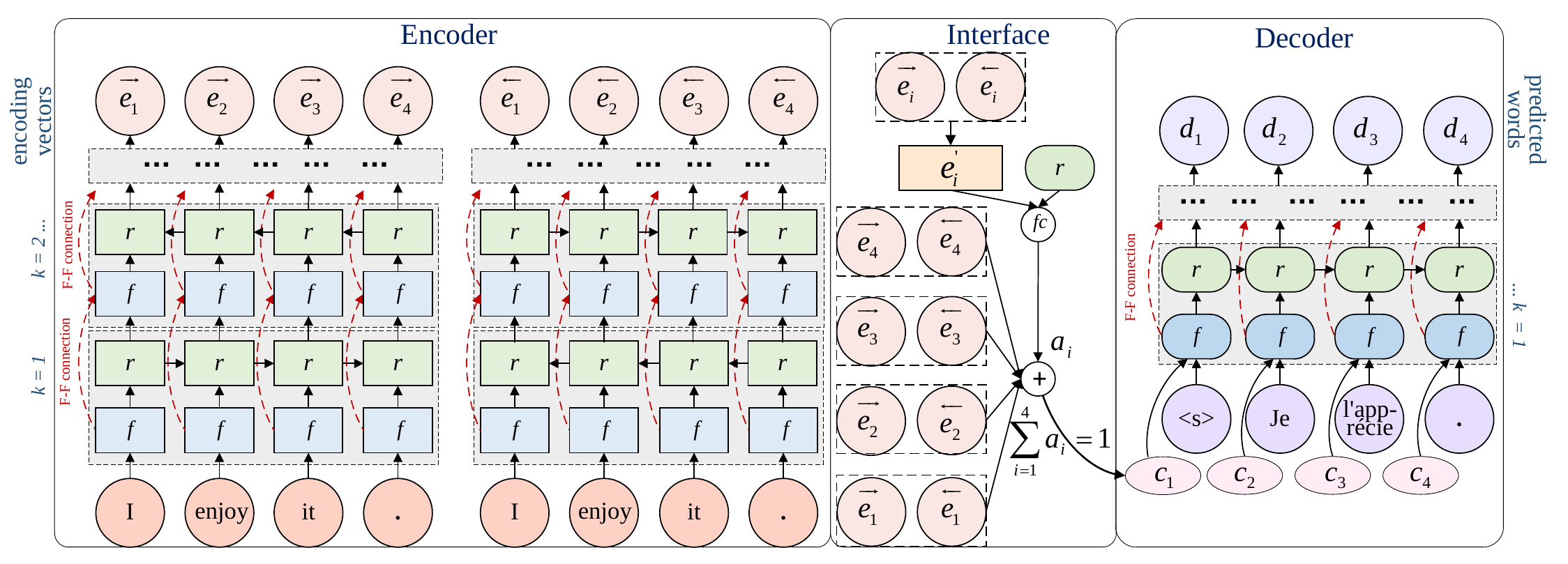}
\caption{The network. It includes three parts from left to right: encoder part (P-E), interface (P-I) and decoder part (P-D). We only
show the topology of Deep-Att as an example.  ``f'' and ``r'' blocks correspond to the feed-forward part and the subsequent
LSTM computation.    The F-F connections are denoted by dashed red lines.}
\label{fig:NMT-Model-Topology}
\end{center}
\end{figure*}

Our entire deep neural network is shown in Fig.~\ref{fig:NMT-Model-Topology}. This topology can be divided into three parts:
the encoder part (P-E) on the left, the decoder part (P-D) on the right and the interface between these two parts (P-I)
which extracts the representation of the source sequence. We have two instantiations of this topology: Deep-ED and Deep-Att, which  correspond
to the extension of the encoder-decoder network and the attention network respectively. Our main innovation is the novel
scheme for connecting adjacent recurrent layers. We will start with the basic RNN model for the sake of clarity.
\\
\noindent\textbf{Recurrent layer:} When an input sequence $\{x_1,\dots,x_m\}$ is given to a recurrent layer, the output $h_t$ at each time
step $t$ can be computed as (see Fig.~\ref{fig:NMT-Model-RNN-FF} (a))
\begin{eqnarray}
\label{eqn:NMT-Model-Recurrent}
  h_t &=& \sigma(W_{f}x_{t}+W_{r}h_{t-1}) \nonumber \\
      &=& \mathrm{RNN}~(W_f x_t, h_{t-1}) \nonumber \\
      &=& \mathrm{RNN}~(f_t, h_{t-1}) ,
\end{eqnarray}
where the bias parameter is not included for simplicity. We use a red circle and a blue empty square to denote an input and a hidden
state. A blue square with a ``-'' denotes the previous hidden state. A dotted line means that the
hidden state is used recurrently.
This computation can be equivalently split into two consecutive steps:
\begin{itemize}
  \item Feed-Forward computation: $f_t=W_{f} x_t$. Left part in Fig.~\ref{fig:NMT-Model-RNN-FF} (b). ``f'' block.
  \item Recurrent computation: $\mathrm{RNN}~(f_t, h_{t-1})$. Right part and the sum operation (+) followed by activation in
      Fig.~\ref{fig:NMT-Model-RNN-FF} (b). ``r'' block.
\end{itemize}
%

For a deep topology with stacked recurrent layers, the input of each block ``f'' at recurrent layer $k$ (denoted by $f^k$) is usually the
output of block ``r'' at its previous recurrent layer $k-1$ (\mbox{denoted} by $h^{k-1}$). In our work, we add \textbf{fast-forward
connections} (F-F connections) which connect two feed-forward computation blocks ``f'' of adjacent recurrent layers. It means that each block
``f'' at recurrent layer $k$ takes both the outputs of block ``f'' and block ``r'' at its previous layer as input
(Fig.~\ref{fig:NMT-Model-RNN-FF}~(c)). F-F connections are denoted by dashed red lines in Fig.~\ref{fig:NMT-Model-RNN-FF}~(c) and
Fig.~\ref{fig:NMT-Model-Topology}. The path of F-F connections contains neither nonlinear activations nor recurrent computation. It provides a
fast path for information to propagate, so we call this path fast-forward connections.

Additionally, in order to learn more temporal dependencies, the sequences can be processed in \mbox{different} directions at each pair of
adjacent recurrent layers. This is quantitatively expressed in Eq.~\ref{eqn:NMT-Model-LSTM}:
\begin{eqnarray}
\label{eqn:NMT-Model-LSTM}
f^k_t &=& W_{f}^k \cdot [f^{k-1}_t, h^{k-1}_t],  ~~~k>1 \nonumber\\
f^k_t &=& W_{f}^k x_t ~~~~~~~~~~~~~~~~~~~~~~~~k=1 \nonumber\\
h^k_t &=& \mathrm{RNN}^k~(f^k_t, h^k_{t+(-1)^k})
\end{eqnarray}
The opposite directions are marked by the direction term $(-1)^k$. At the first recurrent layer, the block ``f'' takes $x_t$ as the input.
$[~,~]$ denotes the concatenation of  vectors. This is shown in Fig.~\ref{fig:NMT-Model-RNN-FF} (c). The two changes are summarized here:
\begin{itemize}
\item We add a connection between $f_t^k$ and $f_t^{k-1}$. Without $f_t^{k-1}$, our model will be reduced to the traditional stacked model.
\item We alternate the RNN direction at different layers $k$ with the direction term $(-1)^k$. If we fix the direction term to
    $-1$, all layers work in the forward direction.
\end{itemize}
\noindent\textbf{LSTM layer:} In our experiments, instead of an RNN, a specific type of recurrent layer called LSTM
~\cite{Hochreiter-Schmidhuber-NC1997,Graves-Schmidhuber-TPAMI2009}  is used. The LSTM is structurally more complex than the basic RNN
in Eq.~\ref{eqn:NMT-Model-Recurrent}.   We define the computation of the LSTM as a function which maps the input $f$ and its state-output
 pair $(h, s)$ at the previous time step to the current state-output pair. The exact computations for $(h_t,
 s_t)=\mathrm{LSTM}(f_t, h_{t-1}, s_{t-1})$ are the following:
\begin{eqnarray}
\label{eqn:LSTM-short}
 [z, z_\rho, z_\phi, z_\pi] &=& f_t + W_r h_{t-1} \nonumber\\
 s_t &=& \sigma_i(z) \circ \sigma_g(z_\rho + s_{t-1} \circ \theta_\rho)+ \nonumber\\
  && \sigma_g(z_\phi + s_{t-1} \circ \theta_\phi) \circ s_{t-1} \nonumber\\
 h_t &=& \sigma_o(s_t) \circ \sigma_g(z_\pi + s_t \circ \theta_\pi)
\end{eqnarray}
where $[z, z_\rho, z_\phi, z_\pi]$ is the concatenation of four vectors of equal size,
$\circ$ means element-wise multiplication, $\sigma_i$ is the input activation function,
$\sigma_o$ is the output activation function, $\sigma_g$ is the activation function for gates,
and $W_r$, $\theta_\rho$, $\theta_\phi$, and $\theta_\pi$ are the parameters of the LSTM. It is slightly different from the standard notation
in that we do not  have a matrix to multiply with the input $f$ in our notation.

With this notation, we can write down the computations for our deep bi-directional LSTM model with F-F connections:
\begin{align}
\label{eqn:NMT-Model-LSTM-Stack}
%
f^k_t &= W_f^k\cdot [f^{k-1}_t, h^{k-1}_t], \;\;  k>1 \nonumber\\
f^k_t &= W_f^k x_t, \;\;\;\;\;\;\;\;\;\;\;\;\;\;\;\;\;\;\;\; k=1 \nonumber\\
(h^k_t, s^k_t) &= \mathrm{LSTM}^k\left(f^k_t, h^k_{t+(-1)^k}, s^k_{t+(-1)^k}\right)
\end{align}
where $x_t$ is the input to the deep bi-directional \mbox{LSTM} model. For the encoder, $x_t$ is the embedding of the
$t^{th}$ word in the source sentence. For the decoder $x_t$ is the concatenation of the embedding of the $t^{th}$ word in the target
sentence and the encoder representation for step $t$.

In our final model， two additional operations are used with Eq.~\ref{eqn:NMT-Model-LSTM-Stack}, which is  shown in
Eq.~\ref{eqn:NMT-Model-LSTM-Stack-Mod}. $\mathrm{Half}(f)$ denotes the first half of the elements of $f$, and
$\mathrm{Dr}(h)$ is the dropout operation ~\cite{Hinton-Salakhutdinov-Arxiv2012} which randomly sets an element of $h$ to zero with a
certain probability. The use of $\mathrm{Half}(\cdot)$ is to reduce the parameter size and does not affect the performance. We
observed noticeable performance degradation when using only the first third of the elements of ``f''.
\begin{equation}
\label{eqn:NMT-Model-LSTM-Stack-Mod}
f^k_t = W_f^k\cdot [\mathrm{Half}(f^{k-1}_t), \mathrm{Dr}(h^{k-1}_t)], \;\;  k>1
\end{equation}

With the F-F connections, we build a fast channel  to propagate the gradients in the deep topology.  F-F connections can accelerate
the model convergence and while improving the performance. A similar idea was also used in ~\cite{He-Sun-ARXIV2015,Zhou-Xu-ACL2015}.

\noindent\textbf{Encoder:} The LSTM layers are stacked following Eq.~\ref{eqn:NMT-Model-LSTM-Stack}. We call this type of encoder
\textbf{interleaved bi-directional encoder}. In addition, there are two similar columns ($a_1$ and $a_2$) in the encoder part. Each
column consists of $n_e$ stacked LSTM layers. There is no connection between the two columns. The first layers of the two columns
process the word representations  of the source sequence in different directions. At the last LSTM layers, there are two groups of
vectors representing the source sequence. The group size is the same as the length of the input sequence.

\noindent\textbf{Interface:} Prior encoder-decoder models and attention models are different in their method of
extracting the representations of the source sequences. In our work, as a consequence of the introduced F-F connections, we have $4$
output vectors ($h_t^{n_e}$ and $f_t^{n_e}$ of both columns). The representations are modified for both Deep-ED and Deep-Att.

For Deep-ED, $e_t$ is static and consists of four \mbox{parts}. 1: The last time step output $h_m^{n_e}$ of the first column. 2: Max-operation
$\mathrm{Max}(\cdot)$ of $h_t^{n_e}$ at \mbox{all} time steps of the second column, denoted by $\mathrm{Max}(h_t^{n_e, a_2})$.
$\mathrm{Max}(\cdot)$ denotes  obtaining the maximal value for each dimension over $t$. 3: $\mathrm{Max}(f_t^{n_e, a_1})$. 4:
$\mathrm{Max}(f_t^{n_e, a_2})$. The max-operation and last time step state extraction provide complimentary information but do not
affect the performance much. $e_t$ is  used as the final representation $c_t$.

For Deep-Att, we do not need the above two operations. We only concatenate the $4$ output vectors at each time step to obtain $e_t$,
and a soft attention mechanism~\cite{Bahdanau-Bengio-ICLR2015} is used to calculate the final representation $c_t$ from $e_t$. $e_t$ is
summarized as:
\begin{align}
\small
\label{eqn:lstm-rep}
&\textrm{Deep-ED:}~~~e_t  \nonumber\\
&~~~[h_m^{n_e, a_1}, \mathrm{Max}(h_t^{n_e, a_2}), \mathrm{Max}(f_t^{n_e, a_1}), \mathrm{Max}(f_t^{n_e, a_2})] \nonumber\\
&\textrm{Deep-Att:} ~~~e_t \nonumber\\
&~~~ [h_t^{n_e, a_1}, h_t^{n_e, a_2}, f_t^{n_e, a_1}, f_t^{n_e, a_2}]
\end{align}
Note that the vector dimensionality of $f$ is four times larger than that of $h$ (see Eq.~\ref{eqn:LSTM-short}). $c_t$ is summarized
as:
\begin{align}
\small
\label{eqn:lstm-rep2}
&\textrm{Deep-ED:} ~~~c_t=e_t, ~~~ \mathrm{(const)} \nonumber\\
&\textrm{Deep-Att:}~~~ c_t=\sum_{t'=1}^m \alpha_{t,t'}  W_p e_{t'} ~~~~~~~~~~
\end{align}
$\alpha_{t,t'}$ is the normalized attention weight computed by:
\begin{equation}
\alpha_{t,t'}=\frac{\exp(a(W_p e_{t'}, h_{t-1}^{1,dec}))}{\sum_{t''}\exp(a(W_p e_{t''}, h_{t-1}^{1,dec}))}
\end{equation}
$h_{t-1}^{1, dec}$ is the first hidden layer output in the decoding part. $a(\cdot)$ is an alignment model described
in~\cite{Bahdanau-Bengio-ICLR2015}.  For Deep-Att, in order to reduce the memory cost, we linearly project (with $W_p$) the
concatenated vector $e_t$  to a vector with $1/4$ dimension size, denoted by the (fully connected) block ``fc'' in
Fig.~\ref{fig:NMT-Model-Topology}.

\noindent\textbf{Decoder:}  The decoder follows Eq.~\ref{eqn:NMT-Model-LSTM-Stack} and Eq.~\ref{eqn:NMT-Model-LSTM-Stack-Mod}
with fixed direction term $-1$. At the first layer, we use the following $x_t$:
\begin{equation}
\label{eqn:lstm-context}
x_t=[c_t, y_{t-1}]
\end{equation}
$y_{t-1}$ is the target word embedding at the previous time step and  $y_0$ is zero. There is a single column of $n_d$ stacked LSTM layers. We
also use the F-F connections like those in the encoder and all layers are in the forward direction.   Note that at the last LSTM
layer, we only use $h_t$ to make the prediction with a softmax layer.


Although the network is deep, the training technique is straightforward. We will describe this in the next part. 

\subsection{Training technique}

We  take the parallel data as the only input without using any monolingual data for either word representation pre-training or  language
modeling. Because of the deep bi-directional structure, we do not need to reverse the sequence order as ~\newcite{Sutskever-Le-NIPS2014}.

The deep topology brings difficulties for the model training, especially when  first order methods such as stochastic gradient descent (SGD)
~\cite{Lecun-Haffner-IEEE1998} are used. The parameters should be properly initialized and the converging process can be slow. We tried
several  optimization techniques such as AdaDelta ~\cite{Zeiler-ARXIV2012}, RMSProp ~\cite{Tieleman-Hinton-Lecture2012}  and Adam
~\cite{Kingma-Ba-ICLR2015}. We found that all of them were able to speed up the process a lot compared to simple SGD while no significant
performance difference  was observed \mbox{among} them. In this work, we chose Adam for model training and do not present a detailed
comparison with other optimization methods.

Dropout ~\cite{Hinton-Salakhutdinov-Arxiv2012} is also used to avoid over-fitting. It is utilized on the LSTM nodes $h_{t}^k$ (See
Eq.~\ref{eqn:NMT-Model-LSTM-Stack}) with a ratio of $p_{d}$ for both the encoder and decoder.

During the whole model training process, we keep all hyper parameters fixed without any intermediate interruption. The hyper parameters are
selected according to the performance on the development set. For such a deep and large network, it is not easy to determine the tuning
strategy and this will be considered in future work.

\subsection{Generation}

We use the common left-to-right beam-search method for sequence generation. At each time step $t$, the word $y_t$ can be predicted by:
\begin{equation}
  \hat{y}_t = \mathrm{arg}~\max_{y}~\mathrm{P}(y|\boldsymbol{\hat{y}}_{0:t-1}, \boldsymbol{x}; \boldsymbol{\theta})
  \label{eqn:NMT-Model-gen}
\end{equation}
where $\hat{y}_t$ is the predicted target word. $\boldsymbol{\hat{y}}_{0:t-1}$ is the generated sequence from time step $0$ to $t-1$. We keep
$n_b$ best candidates according to Eq.~\ref{eqn:NMT-Model-gen} at each time step, until the end of sentence mark is generated. The hypotheses
are ranked by the total likelihood of the generated sequence, although normalized likelihood is used in some
works~\cite{Jean-Bengio-ACL2015}.

\section{Experiments}

We evaluate our method mainly on the widely used WMT'14 English-to-French translation task. In order to validate our model on more difficult
language pairs, we also provide results on the WMT'14 English-to-German translation task. Our models are implemented in the PADDLE (PArallel
Distributed Deep LEarning) platform.

\subsection{Data sets}

For both tasks, we use the full WMT'14  parallel corpus as our training data. The detailed data sets are listed below:
\begin{itemize}
\item English-to-French: Europarl v7, Common Crawl, UN, News Commentary, Gigaword
\item English-to-German: Europarl v7, Common Crawl, News Commentary
\end{itemize}
In total, the English-to-French corpus includes $36$ million sentence pairs, and the English-to-German corpus includes $4.5$ million sentence
pairs.
The news-test-2012 and news-test-2013 are concatenated as our development set, and the news-test-2014 is the test set. Our data partition is
consistent with previous works on NMT ~\cite{Luong-Zaremba-ACL2015,Jean-Bengio-ACL2015}  to ensure fair comparison.

For the source language, we select the most frequent $200$K words as the input vocabulary. For the target language we select the most frequent
$80$K French words and the most frequent $160$K German words as the output vocabulary. The full vocabulary of the German corpus is
larger~\cite{Jean-Bengio-ACL2015}, so we select more German words to build the target vocabulary. Out-of-vocabulary words are replaced with
the unknown symbol $\langle \mathrm{unk}\rangle$. For complete comparison to previous work on the English-to-French task, we also show the
results with a smaller vocabulary of $30$K input words and $30$K output words on the sub train set with selected $12$M parallel sequences
~\cite{Schwenk-2014,Sutskever-Le-NIPS2014,Cho-Bengio-EMNLP2014}.

\subsection{Model settings}

We have two models as described above, named Deep-ED and Deep-Att. Both models have  exactly the same configuration and layer size  except the
interface part P-I.

We use $256$ dimensional word embeddings for both the source and target languages.  All LSTM \mbox{layers}, including the $2\times n_e$ layers
in the encoder and the $n_d$ layers in the decoder, have $512$ memory cells. The output layer size is the same as the size of the target
vocabulary. The dimension of $c_t$ is $5120$ and $1280$ for Deep-ED and Deep-Att respectively. For each LSTM layer, the activation functions
for gates, inputs and outputs are $\mathrm{sigmoid}$, $\mathrm{tanh}$, and $\mathrm{tanh}$ respectively.

Our network is narrow on word embeddings and LSTM layers. Note that in previous work ~\cite{Sutskever-Le-NIPS2014,Bahdanau-Bengio-ICLR2015},
$1000$ dimensional word embeddings and $1000$ dimensional \mbox{LSTM} layers are used. We also tried larger scale models but did not obtain
further improvements.

\subsection{Optimization}

Note that each LSTM layer includes two parts as described in Eq.~\ref{eqn:NMT-Model-LSTM}, feed-forward computation and recurrent computation.
\mbox{Since} there are non-linear activations in the recurrent computation, a larger learning rate $l_r=5\times10^{-4}$ is used, while for the
feed-forward computation a smaller learning rate $l_f=4\times10^{-5}$ is used. Word embeddings and the softmax layer also use this learning
rate $l_f$. We refer all the parameters not used for recurrent computation as non-recurrent part of the model.

Because of the large model size, we use strong $L_2$ regularization to constrain the parameter matrix $v$ in the following way:
\begin{equation}
v \leftarrow v - l\cdot(g+r\cdot v)
\end{equation}
Here $r$ is the regularization strength, $l$ is the corresponding learning rate, $g$ stands for the gradients of $v$. The two embedding layers
are not regularized. All the other layers have the same  $r=2$.

The parameters of the recurrent computation part are initialized to zero. All non-recurrent parts are randomly initialized with zero mean and
standard deviation of $0.07$. A detailed guide for setting hyper-parameters can be found in ~\cite{Bengio-ARXIV2012}.

The dropout ratio $p_{d}$ is  $0.1$. In each batch, there are $500\sim800$ sequences in our work. The exact number depends on the sequence
lengths and model size.  We also find that larger batch size results in better convergence although the improvement is not large. However, the
largest batch size is constrained  by the GPU memory. We use $4\sim8$ GPU machines (each has $4$ K40 GPU cards) running for $10$ days to train
the full model with parallelization at the data batch level. It takes nearly $1.5$ days for each pass.

One thing we want to emphasize here is that our deep model is not sensitive to these settings. Small variation does not affect the final
performance.

\subsection{Results}

We evaluate the same way as previous NMT works ~\cite{Sutskever-Le-NIPS2014,Luong-Zaremba-ACL2015,Jean-Bengio-ACL2015}. All reported BLEU
\mbox{scores} are computed with the
\emph{multi-bleu.perl}\footnote{\url{https://github.com/moses-smt/mosesdecoder/blob/master/scripts/generic/multi-bleu.perl}} script which is
also used in the above works. The results are for tokenized and case sensitive evaluation.

\subsubsection{Single models}

\noindent\textbf{English-to-French:} ~First we list our single model results on the English-to-French task in
Tab.~\ref{tab:Results-BLEU-Single}. In the first block we show the results with the \mbox{full} corpus. The previous best single NMT
encoder-decoder model (Enc-Dec) with six layers achieves BLEU=$31.5$ ~\cite{Luong-Zaremba-ACL2015}. From Deep-ED, we obtain the BLEU score of
$36.3$, which outperforms Enc-Dec model by $4.8$ BLEU points. This result is even better than the ensemble result of eight Enc-Dec models,
which is $35.6$ ~\cite{Luong-Zaremba-ACL2015}. This shows that, in addition to the convolutional layers for computer vision, deep topologies
can also work for LSTM layers. For Deep-Att, the performance is further improved to  $37.7$. We also list the previous state-of-the-art
performance from a conventional SMT system ~\cite{Durrani-Heafield-WMT2014} with the BLEU of $37.0$. This is the first time that a single NMT
model trained in an end-to-end form beats the best \mbox{conventional} system on this task.

We also show the results on the smaller data set with $12$M sentence pairs and $30$K vocabulary in the second block. The two attention models,
RNNsearch ~\cite{Bahdanau-Bengio-ICLR2015} and RNNsearch-LV ~\cite{Jean-Bengio-ACL2015}, achieve BLEU scores of $28.5$ and $32.7$
respectively. Note that RNNsearch-LV uses a large output vocabulary of $500$K words based on the standard attention model RNNsearch.  We
obtain  BLEU=$35.9$  which outperforms its corresponding shallow model RNNsearch by $7.4$ BLEU points. The SMT result from
~\cite{Schwenk-2014} is also listed and falls behind our model by $2.6$ BLEU points.

\begin{table}[!ht]
\footnotesize
\begin{center}
\begin{tabular}{|l|c|c|c|}
  \hline
  Methods &  Data  & Voc & BLEU \\
  \hline
  \hline
  Enc-Dec (Luong,2015)  & 36M & 80K & 31.5 \\
  \hline
  SMT (Durrani,2014)    &  36M & Full & 37.0 \\
  \hline
  Deep-ED (Ours)    & 36M & 80K & 36.3 \\
  Deep-Att (Ours)  & 36M & 80K & \textbf{37.7} \\
  \hline
  \hline
  RNNsearch (Bahdanau,2014)  & 12M & 30K & 28.5 \\
  RNNsearch-LV (Jean,2015)  & 12M & 500K & 32.7 \\
  \hline
  SMT (Schwenk,2014) & 12M & Full & 33.3 \\
  \hline
  Deep-Att (Ours)  & 12M & 30K & 35.9  \\
  \hline
\end{tabular}
\end{center}
\caption{\label{tab:Results-BLEU-Single} English-to-French task: BLEU scores of single neural models. We also list the conventional SMT system
for comparison. }
\end{table}

Moreover, during the generation process, we obtained the best BLEU score with beam size $=3$ (when the beam size is $2$, there is only a $0.1$
difference in BLEU score). This is different from other \mbox{works} listed in Tab.~\ref{tab:Results-BLEU-Single}, where the beam size is $12$
~\cite{Jean-Bengio-ACL2015,Sutskever-Le-NIPS2014}. We attribute this \mbox{difference} to the improved model performance, where the ground
truth generally exists in the top hypothesis. Consequently, with the much smaller  beam size, the generation efficiency is significantly
improved.

Next we  list the effect of the novel F-F connections in our Deep-Att model of shallow topology in Tab.~\ref{tab:Results-BLEU-SingleCheck}.
When $n_e=1$ and $n_d=1$, the BLEU scores are $31.2$ without F-F and $32.3$ with F-F. Note that the model without F-F is exactly the standard
attention model~\cite{Bahdanau-Bengio-ICLR2015}. Since there is only a single layer, the use of F-F \mbox{connections} means that at the
interface part we include $f_t$ into the representation (see Eq.~\ref{eqn:lstm-rep}).  We find F-F connections bring an improvement of $1.1$
in BLEU. After we increase our model depth to $n_e=2$ and $n_d=2$, the improvement is enlarged to $1.4$.  When the model is trained with
larger depth without F-F connections, we find that the parameter exploding problem~\cite{Bengio-Frasconi-Trans1994} happens so frequently that
we could not finish training. This suggests that F-F connections provide a fast way for gradient propagation.
\begin{table}[!ht]
\footnotesize
\begin{center}
\begin{tabular}{|l|c|c|c|c|}
  \hline
  Models & F-F &$n_e$ & $n_d$ & BLEU \\
  \hline
  Deep-Att  & No  & 1  & 1 & 31.2 \\
  Deep-Att  & Yes & 1  & 1 & 32.3 \\
  \hline
  Deep-Att  & No  & 2  & 2 & 33.3 \\
  Deep-Att  & Yes & 2  & 2 & 34.7 \\
  \hline
\end{tabular}
\end{center}
\caption{\label{tab:Results-BLEU-SingleCheck} The effect of F-F. We list the BLEU scores  of Deep-Att with and without F-F. Because of the
parameter exploding problem, we can not list the model performance of larger depth without F-F. For $n_e=1$ and $n_d=1$, F-F connections only
contribute to the representation at interface part (see Eq.~\ref{eqn:lstm-rep}).}
\end{table}

Removing F-F connections also reduces the corresponding model size. In order to figure out the effect of F-F comparing models with the same
parameter size, we increase the LSTM layer width of Deep-Att without F-F. In Tab.~\ref{tab:Results-BLEU-SizeCheck} we show that, after using a
two times larger LSTM layer width of $1024$, we can only obtain a BLEU score of $33.8$, which is still worse than the corresponding Deep-Att
with F-F.
\begin{table}[!ht]
\footnotesize
\begin{center}
\begin{tabular}{|l|c|c|c|c|c|}
  \hline
  Models & F-F &$n_e$ & $n_d$ & width & BLEU \\
  \hline
  Deep-Att  & No  & 2  & 2 & 512 & 33.3 \\
  Deep-Att  & No  & 2  & 2 & 1024 & 33.8 \\
  \hline
  Deep-Att  & Yes & 2  & 2 & 512 & 34.7 \\
  \hline
\end{tabular}
\end{center}
\caption{\label{tab:Results-BLEU-SizeCheck} BLEU scores with different LSTM layer width in Deep-Att. After using two times larger LSTM layer
width of $1024$, we can only obtain BLEU score of $33.8$. It is still behind the corresponding Deep-Att with F-F.}
\end{table}

We also notice that the interleaved bi-directional encoder starts to work when the encoder depth is larger than $1$. The effect of the
interleaved bi-directional encoder is shown in Tab.~\ref{tab:Results-BLEU-BiCheck}. For our largest model with $n_e=9$ and $n_d=7$, we
compared the BLEU scores of the interleaved bi-directional encoder and the uni-directional encoder (where all LSTM layers work in forward
direction). We find there is a gap of about $1.5$ points between these two encoders for both Deep-Att and Deep-ED.
\begin{table}[!ht]
\footnotesize
\begin{center}
\begin{tabular}{|l|c|c|c|c|c|}
  \hline
  Models & Encoder &$n_e$ & $n_d$ & BLEU \\
  \hline
  Deep-Att  & Bi  & 9  & 7 & 37.7 \\
  Deep-Att  & Uni & 9  & 7 & 36.2 \\
  \hline
  Deep-ED  & Bi  & 9  & 7 & 36.3 \\
  Deep-ED  & Uni & 9  & 7 & 34.9 \\
  \hline
\end{tabular}
\end{center}
\caption{\label{tab:Results-BLEU-BiCheck} The effect of the interleaved bi-directional encoder. We list the BLEU scores  of our largest
Deep-Att and Deep-ED models. The encoder term Bi denotes that the interleaved bi-directional encoder is used. Uni denotes a model where all
LSTM layers work in forward direction.}
\end{table}

Next we look into the effect of model depth. In Tab.~\ref{tab:Results-BLEU-SingleDepthCheck}, starting from $n_e=1$ and $n_d=1$ and gradually
increasing the model depth, we significantly increase BLEU scores. With $n_e=9$ and $n_d=7$, the best score for Deep-Att is $37.7$. We tried
to increase the LSTM width based on this, but obtained little improvement. As we stated in Sec.\ref{sec:NMT}, the complexity of the encoder
and decoder, which is related to the model depth, is more important than the model size. We also tried a larger depth, but the results started
to get worse. With our topology and training technique, $n_e=9$ and $n_d=7$ is the best depth we can achieve.
\begin{table}[!ht]
\footnotesize
\begin{center}
\begin{tabular}{|l|c|c|c|c|c|}
  \hline
  Models & F-F &$n_e$ & $n_d$ & Col & BLEU \\
  \hline
  Deep-Att  & Yes & 1  & 1 & 2 & 32.3 \\
  Deep-Att  & Yes & 2  & 2 & 2 & 34.7 \\
  Deep-Att  & Yes & 5  & 3 & 2 & 36.0 \\
  Deep-Att  & Yes & 9 &  7 & 2 & 37.7 \\
  Deep-Att  & Yes & 9  & 7 & 1 & 36.6 \\
  \hline
\end{tabular}
\end{center}
\caption{\label{tab:Results-BLEU-SingleDepthCheck} BLEU score of Deep-Att with different model depth. With $n_e=1$ and $n_d=1$, F-F
connections only contribute to the representation at interface part£¬ where $f_t$ is included (see Eq.~\ref{eqn:lstm-rep}).}
\end{table}

The last line in Tab.~\ref{tab:Results-BLEU-SingleDepthCheck} shows the BLEU \mbox{score} of $36.6$ of our deepest model, where only one
encoding column ($\textrm{Col}=1$) is used. We find a $1.1$ BLEU points degradation with a single encoding column. Note that the
uni-directional models in Tab.~\ref{tab:Results-BLEU-BiCheck}  with uni-direction still have two encoding columns.  In order to find out
whether this is caused by the decreased parameter size, we test a wider model with 1024 memory blocks for the LSTM layers. It is shown in
Tab.~\ref{tab:Results-BLEU-SingleColCheck} that there is a minor improvement of only $0.1$. We attribute this to the complementary information
provided by the double encoding column.
\begin{table}[!ht]
\footnotesize
\begin{center}
\begin{tabular}{|l|c|c|c|c|c|c|}
  \hline
  Models & F-F &$n_e$ & $n_d$ & Col & width &BLEU \\
  \hline
  Deep-Att  & Yes & 9 &  7 & 2 & 512 & 37.7 \\
  Deep-Att  & Yes & 9  & 7 & 1 & 512 & 36.6 \\
  Deep-Att  & Yes & 9  & 7 & 1 & 1024 & 36.7 \\
  \hline
\end{tabular}
\end{center}
\caption{\label{tab:Results-BLEU-SingleColCheck} Comparison of encoders with different number of columns and LSTM layer width.}
\end{table}

\textbf{English-to-German:} We also validate our deep topology on the English-to-German task. The English-to-German task is considered a
relatively more \mbox{difficult} task, because of the lower similarity between these two languages. Since the German vocabulary is much larger
than the French vocabulary, we select $160$K most frequent words as the target vocabulary. All the other hyper parameters are exactly the same
as those in the English-to-French task.

We list our single model Deep-Att performance in Tab.~\ref{tab:Results-BLEU-SingleGerman}. Our single model result with BLEU=$20.6$ is similar
to the conventional SMT result of $20.7$~\cite{Buck-Ooyen-2014}. We also outperform the shallow attention models as shown in the first two
lines in Tab.~\ref{tab:Results-BLEU-SingleGerman}. All the results are consistent with those in the English-to-French task.
\begin{table}[!ht]
\footnotesize
\begin{center}
\begin{tabular}{|l|c|c|c|}
  \hline
  Methods &  Data  & Voc & BLEU \\
  \hline
  \hline
  RNNsearch (Jean,2015)  & 4.5M & 50K & 16.5 \\
  RNNsearch-LV (Jean,2015)  & 4.5M & 500K & 16.9 \\
  \hline
  SMT (Buck,2014) & 4.5M & Full & 20.7 \\
  \hline
  Deep-Att (Ours)  & 4.5M & 160K & 20.6  \\
  \hline
\end{tabular}
\end{center}
\caption{\label{tab:Results-BLEU-SingleGerman} English-to-German task: BLEU scores of single neural models. We also list the conventional SMT
system for comparison. }
\end{table}

\subsubsection{Post processing}

Two post processing techniques are used to improve the performance further on the English-to-French task.

First,  three Deep-Att models are built for ensemble results. They are initialized with different random parameters;  in addition, the
training corpus for these models is shuffled with different random seeds. We sum over the predicted probabilities  of the target words and
normalize the final distribution to  generate the next word. It is shown in Tab.~\ref{tab:Results-BLEU-Post} that the model ensemble can
improve the performance further to $38.9$. In ~\newcite{Luong-Zaremba-ACL2015} and ~\newcite{Jean-Bengio-ACL2015} there are eight models for
the best scores, but we only use three models and we do not obtain further gain from more models.

\begin{table}[!ht]
\footnotesize
\begin{center}
\begin{tabular}{|l|l|c|c|c|}
  \hline
  Methods & Model & Data & Voc & BLEU \\
  \hline
  \hline
  Deep-ED & Single  & 36M & 80K & 36.3 \\
  Deep-Att & Single & 36M & 80K & 37.7 \\
  \hline
  Deep-Att & Single+PosUnk  & 36M & 80K & 39.2 \\
  Deep-Att & Ensemble  & 36M & 80K & 38.9 \\
  Deep-Att & Ensemble+PosUnk  & 36M & 80K & \textbf{40.4} \\
  \hline
  SMT   & Durrani, 2014  & 36M & Full & 37.0 \\
  Enc-Dec & Ensemble+PosUnk & 36M & 80K & 37.5 \\
  \hline
\end{tabular}
\end{center}
\caption{\label{tab:Results-BLEU-Post} BLEU scores of different models. The first two blocks are our results of  two single models and models
with post processing. In the last block we list two baselines of the best conventional SMT system and NMT system.}
\end{table}

Second,  we recover the unknown words in the generated sequences with the Positional Unknown (\mbox{PosUnk}) model introduced in
~\cite{Luong-Zaremba-ACL2015}. The full parallel corpus is used to obtain the word mappings ~\cite{Liang-Klein-NAACL06}. We find this method
provides an additional $1.5$  BLEU points, which is consistent  with the conclusion in ~\newcite{Luong-Zaremba-ACL2015}.  We obtain the new
BLEU score of $39.2$ with a single Deep-Att model.  For the ensemble models of Deep-Att, the BLEU score rises to $40.4$. In the last two
lines, we list the conventional SMT model ~\cite{Durrani-Heafield-WMT2014} and the previous best neural models based system Enc-Dec
~\cite{Luong-Zaremba-ACL2015} for comparison. We find our best score outperforms the previous best score by nearly $3$ points.

\subsection{Analysis}

\subsubsection{Length}
\begin{figure}[!ht]
\begin{center}
\includegraphics[angle=0,width=0.8\columnwidth]{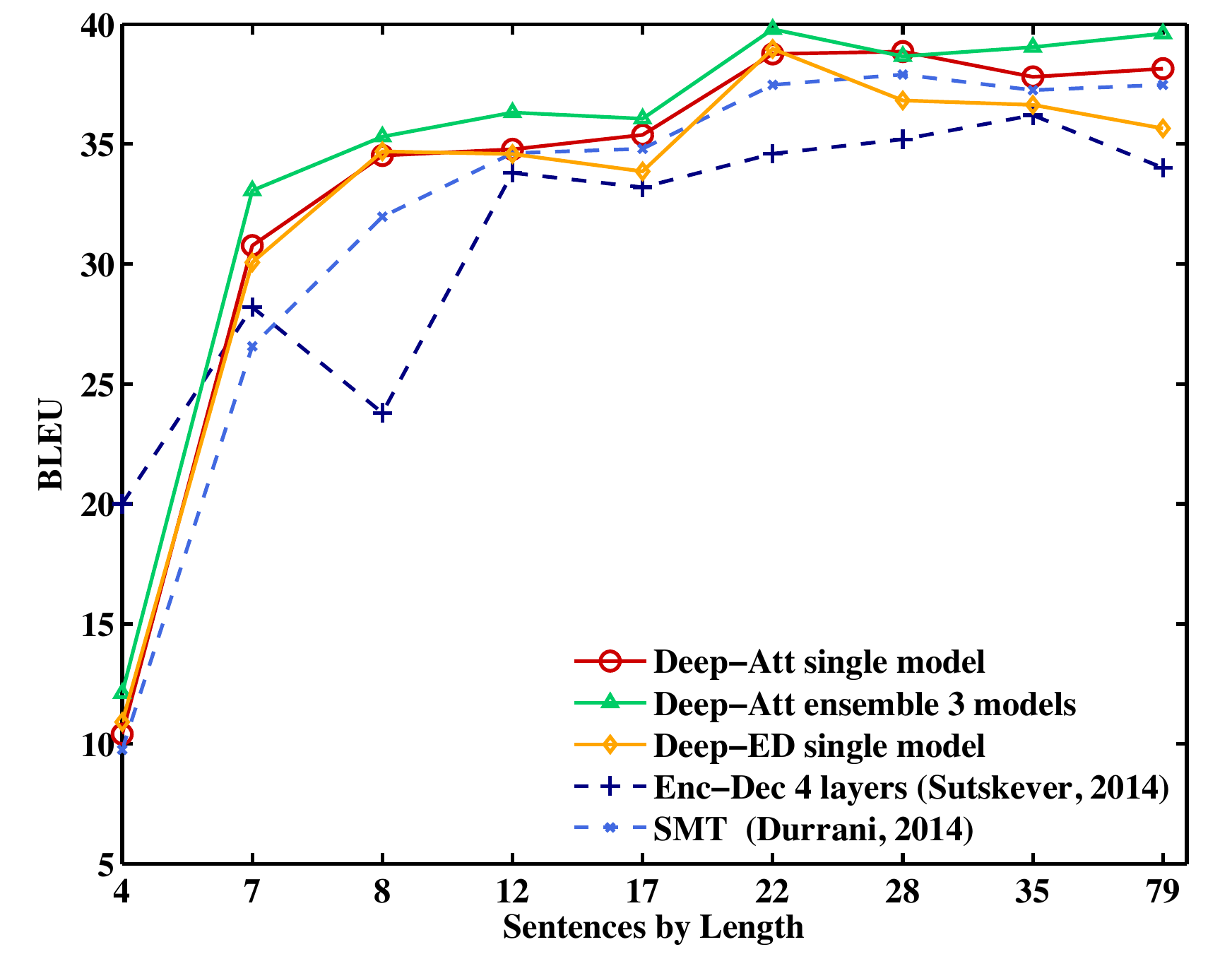}
\caption{BLEU scores vs. source sequence length. Five lines are our Deep-Att single model, Deep-Att ensemble model, our Deep-ED model,
previous Enc-Dec model with four layers and SMT model.}
\label{fig:NMT-Analysis-Length}
\end{center}
\end{figure}

On the English-to-French task, we analyze the effect of the source sentence length on our models as shown in
Fig.~\ref{fig:NMT-Analysis-Length}. Here we show five curves: our Deep-Att single model, our Deep-Att ensemble model, our Deep-ED model, a
previously proposed Enc-Dec model with four layers ~\cite{Sutskever-Le-NIPS2014} and an SMT model ~\cite{Durrani-Heafield-WMT2014}.
We find our Deep-Att model works better than the previous two \mbox{models} (Enc-Dec and SMT) on nearly all sentence lengths. It is also shown
that for very long sequences with length over $70$ words, the performance of our Deep-Att does not degrade, when compared to another NMT model
Enc-Dec. Our Deep-ED also has much better performance than the shallow Enc-Dec model on nearly all lengths, although for long sequences it
degrades and starts to fall behind Deep-Att.

\subsubsection{Unknown words}
Next we look into the detail of the effect of unknown words on the English-to-French task. We select the subset without unknown words on
target sentences from the original test set. There are $1705$ such sentences (56.8\%). We compute the BLEU scores on this subset and the
results are shown in Tab.~\ref{tab:Results-BLEU-SubsetWOunk}. We also list the results from SMT model ~\cite{Durrani-Heafield-WMT2014} as a
comparison.

We find that the BLEU score of Deep-Att on this subset rises to $40.3$, which has a gap of $2.6$ with the score $37.7$ on the full test set.
On this subset, the SMT model achieves  $37.5$,  which is similar to its score $37.0$ on the full test set. This suggests that the
\mbox{difficulty} on this subset is not much different from that on the full set. We therefore attribute the larger gap for Deep-att to the
existence of unknown words. We also compute the BLEU score on the subset of the ensemble model and obtain $41.4$. As a reference related to
human performance, in  ~\newcite{Sutskever-Le-NIPS2014}, it has been tested  that the BLEU score of oracle re-scoring the LIUM $1000$-best
results ~\cite{Schwenk-2014} is $45$.
\begin{table}[!t]
\footnotesize
\begin{center}
\begin{tabular}{|l|l|c|c|}
  \hline
  Model & Test set & Ratio(\%) &  BLEU \\
  \hline
  \hline
  Deep-Att & Full & 100.0  & 37.7 \\
  Ensemble & Full & 100.0  & 38.9 \\
  SMT(Durrani)     & Full & 100.0  & 37.0 \\
  \hline
  Deep-Att & Subset & 56.8  & 40.3 \\
  Ensemble & Subset & 56.8  & 41.4 \\
  SMT(Durrani)     & Subset & 56.8  & 37.5 \\
  \hline
\end{tabular}
\end{center}
\caption{\label{tab:Results-BLEU-SubsetWOunk} BLEU scores of the subset of the test set without considering unknown words. }
\end{table}
\begin{figure}[!t]
\begin{center}
\includegraphics[angle=0,width=0.9\columnwidth]{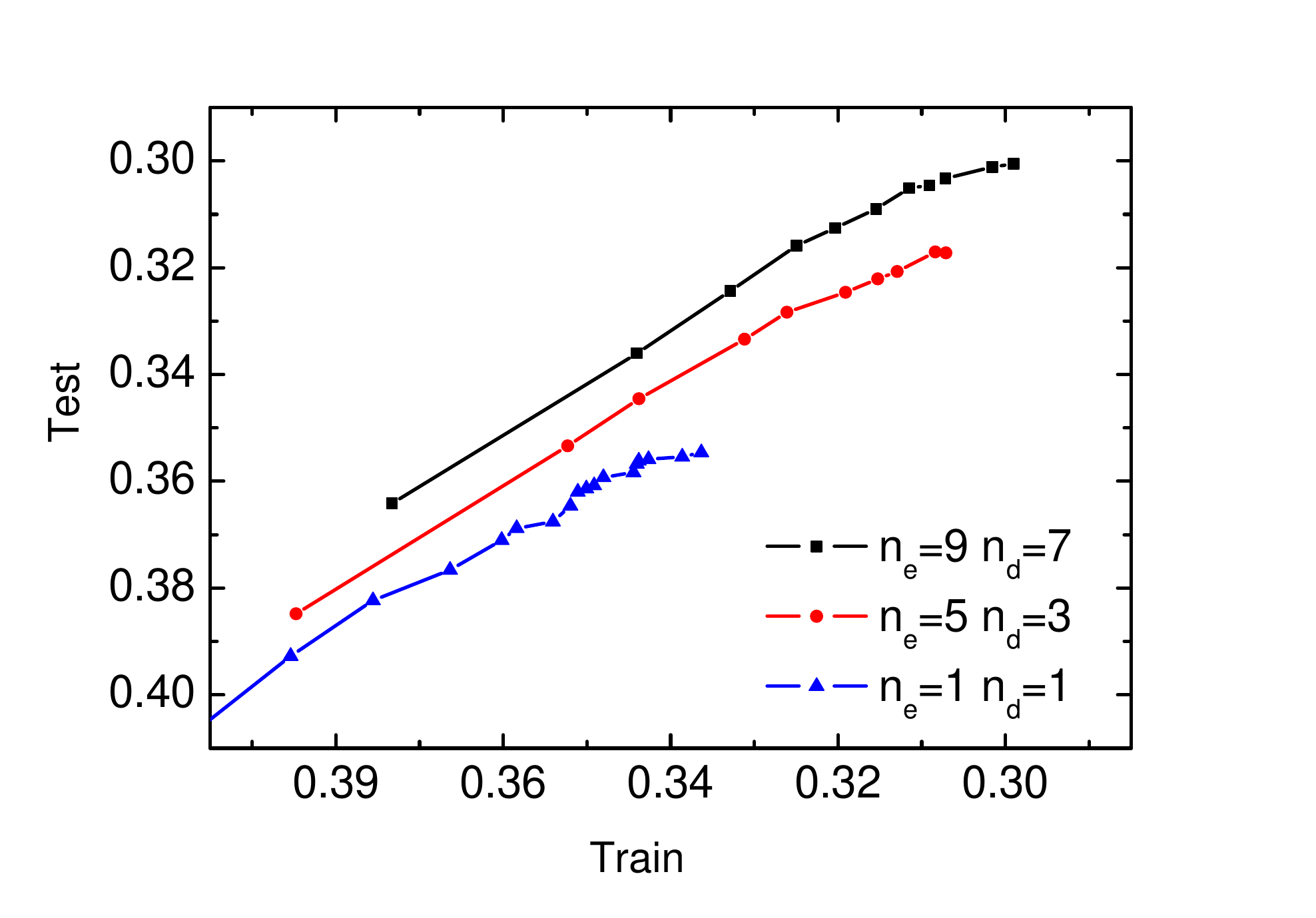}
\caption{Token error rate on train set vs. test set. Square: Deep-Att ($n_e=9$, $n_d=7$). Circle: Deep-Att ($n_e=5$, $n_d=3$). Triagle:
Deep-Att ($n_e=1$, $n_d=1$). }
\label{fig:NMT-Analysis-Overfitting}
\end{center}
\end{figure}
\subsubsection{Over-fitting}

Deep models have more parameters, and thus have a stronger ability to fit the large data set. However, our experimental results suggest that
deep models are less prone to the problem of over-fitting.

In Fig.~\ref{fig:NMT-Analysis-Overfitting}, we show three results from models with a different depth on the English-to-French task. These
three models are evaluated by token error rate, which is defined as the ratio of incorrectly predicted words in the whole target sequence with
correct historical    input.  The curve with square marks corresponds to Deep-Att with $n_e=9$ and  $n_d=7$. The curve with circle marks
corresponds to $n_e=5$ and $n_d=3$. The curve with triangle marks corresponds to $n_e=1$ and $n_d=1$.  We find that the deep model has better
performance on the test set when the token error rate is the same as that of the shallow models on the training set.  This shows that, with
decreased token error rate, the deep model is more advantageous in avoiding the over-fitting phenomenon. We only plot the early training stage
curves because, during the late training stage, the curves are not smooth.

\section{Conclusion}

With the introduction of fast-forward  connections to the deep \mbox{LSTM} network, we build a fast path with neither non-linear
transformations nor recurrent computation to propagate the gradients from the top to the deep bottom. On this path, gradients decay much
slower compared to the  standard deep network. This enables us to build the deep topology of NMT models.

We trained NMT models with depth of $16$ including $25$ LSTM layers and evaluated them mainly on the WMT'14 English-to-French translation
task. This is the deepest topology that has been investigated in the NMT area on this task. We showed that our Deep-Att exhibits $6.2$ BLEU
points \mbox{improvement} over the previous best single model, achieving a $37.7$ BLEU score. This single end-to-end NMT model outperforms the
best conventional SMT system ~\mbox{\cite{Durrani-Heafield-WMT2014}} and achieves a state-of-the-art performance. After utilizing  unknown
word processing and model ensemble of three models, we obtained a BLEU score  of $40.4$, an improvement of  $2.9$ BLEU points over the
previous best result. When evaluated on the subset of the test corpus without unknown words, our model achieves $41.4$. Our model is also
validated on the more difficult English-to-German task.

Our model is also efficient in sequence generation. The best results from both a single model and model ensemble are obtained with a beam size
of $3$, much smaller than previous NMT systems where beam size is about $12$ ~\cite{Jean-Bengio-ACL2015,Sutskever-Le-NIPS2014}.   From our
analysis, we find that deep \mbox{models} are more advantageous for learning for long sequences and that the deep topology is resistant to the
over-fitting problem.

We tried deeper models and did not obtain further improvements with our current topology and training techniques. However, the depth of $16$
is not very deep compared to the models in computer vision ~\cite{He-Sun-ARXIV2015}. We believe we can benefit from deeper models, with new
designs of topologies and training techniques, which remain as our future work.

\bibliographystyle{acl2012}
\bibliography{NMT2015V10}

\end{document}